\documentclass[preprint,12pt]{elsarticle}
\usepackage{amsfonts}  
\usepackage{amsmath}
\usepackage{xcolor}
\usepackage{amsthm}
\usepackage{multirow}
\usepackage{booktabs}
\usepackage{graphicx}
\usepackage{subcaption}
\usepackage{enumitem}
\usepackage{float}

\journal{Journal of Manufacturing Systems}

\begin{document}

\begin{frontmatter}

\title{Robust-MBDL: A Robust Multi-branch Deep
Learning Based Model for Remaining Useful Life
Prediction and Operating Condition Identification of Rotating Machines}

 \author[1]{Khoa~Tran}
 \author[2]{Hai-Canh~Vu\corref{cor1}}
 \author[3]{Lam~Pham}
 \author[4]{Nassim Boudaoud}
\cortext[cor1]{Corresponding author: canh.vuhai@vlu.edu.vn}
 \affiliation[1]{organization={University of Science and Technology - The University of Danang},
     addressline={54 Nguyen Luong Bang, Hoa Khanh Bac, Lien chieu, Da Nang}, 
     city={Danang},
      citysep={}, 
    postcode={550000}, 
    country={Vietnam}}
\affiliation[2]{organization={Laboratory for Applied and Industrial Mathematics, Institute for Computational Science and Artificial Intelligence, Van Lang University\\ Faculty of Mechanical - Electrical and Computer Engineering, School of Technology, Van Lang University},
     city={Ho Chi Minh City},
     postcode={70000}, 
     country={Vietnam}}
     
 \affiliation[3]{organization={AIT Austrian Institute of Technology GmbH},
     addressline={Giefinggasse 4, 1210 Wien}, 
     city={Vienna},
     postcode={1210}, 
     country={Austria}}

 \affiliation[4]{organization={Roberval Laboratory, Department of Mechanical Engineering, University of Technology of Compiègne},
     addressline={Rue Roger Couttolenc}, 
     city={Compiègne},
     postcode={60200}, 
     country={France}}
 
\begin{abstract}
In this paper, a Robust Multi-branch Deep learning-based system for remaining useful life (RUL) prediction and Operating Condition (OC) identification of rotating machines is proposed. 
In particular, the proposed system comprises main components: (1) an LSTM-Autoencoder to denoise the vibration data; (2) a feature extraction to generate time-domain, frequency-domain, and time-frequency based features from the denoised data; (3) a novel and robust multi-branch deep learning network architecture to exploit the multiple features.
The performance of our proposed system was evaluated and compared to the state-of-the-art systems on two benchmark datasets of XJTU-SY and PRONOSTIA. 
The experimental results prove that our proposed system outperforms the state-of-the-art systems and presents potential for real-life applications on bearing machines. 

\end{abstract}
\end{frontmatter}

\section{Introduction}
Accurately estimating the Remaining Useful Life (RUL) plays a pivotal role in predictive maintenance for rotating machines. The prediction of RUL has garnered significant attention from both academic researchers and industry professionals. This is because accurately predicting RUL can significantly enhance the effectiveness of predictive maintenance, leading to increased machine reliability and reduced incidences of failures and associated repair costs.

Existing RUL prediction models generally fall within two primary categories: the model-based and data-driven approaches \cite{ferreira2022remaining}. The model-based approach relies on a certain level of physical knowledge about machine degradation to predict RUL, such as employing theories of the Paris law for bearing defect growth \cite{li2000stochastic} and reliability laws \cite{zhang2021prognostics, behzad2017prognostics, zhang2022wiener}. However, integrating such physical knowledge into models can be challenging, especially concerning complex machinery where such insights might not always be readily available.

The advent of Industrial Internet of Things (IIoT) technologies has facilitated the accumulation of extensive data (evidenced by benchmark datasets for RUL detection, e.g.,\cite{wang2018hybrid},\cite{nectoux2012pronostia}). This influx of data has significantly bolstered the application of the data-driven approach for RUL detection. Unlike model-based methods, the data-driven approach primarily relies on collected data, enabling its application to complex machines/systems without a prerequisite for extensive physical knowledge. 

Machine Learning (ML) is a popular data-driven approach that has been extensively used in predicting the Remaining Useful Life (RUL) of rotating machines. Several studies, including \cite{singh2020systematic}\cite{wu2017comparative}\cite{luo2020hybrid}\cite{sharp2018survey} \cite{liu2019data}, have employed well-known ML models such as Linear Regression (LR), Random Forest (RF), and Support Vector Machines (SVM) to forecast RUL. However, these methods have some significant drawbacks, such as suboptimal performance due to inflexible mathematical formulas and time-consuming computations for big input data. Therefore, there has been a significant shift towards Deep Learning (DL) in preference to traditional ML techniques.

Several deep learning models with simple neural network layers have been proposed for predicting the Remaining Useful Life (RUL) of a machine. One popular model is the Bidirectional Long Short-Term Memory (Bi-LSTM), introduced by Huang et al. in 2019 \cite{huang2019bidirectional}. This model consists of two Bi-LSTM blocks, fully connected layers, and a linear regression layer. The unique feature of the Bi-LSTM components is that they can capture both past and future information simultaneously, which helps to improve the accuracy of RUL estimation. Another recent innovation in this field is the Self-Attention Augmented Convolutional Gated Recurrent Unit Network (SACGNet), which was introduced by Xu et al. in 2022 \cite{xu2022sacgnet}. The research showed that incorporating self-attention mechanisms helps the model focus on critical information, thus enhancing the performance of the Gated Recurrent Unit (GRU) in predicting RUL. SACGNet was compared to other models, such as the Convolutional Neural Network (CNN), Long Short-Term Memory (LSTM), GRU, and Recurrent Neural Network (RNN), using both the PRONOSTIA \cite{nectoux2012pronostia} and XJTU \cite{wang2018hybrid} datasets.

In order to enhance the performance of individual DL models and extract pertinent features more effectively, Al-Dulaimi et al. \cite{al2019hybrid} proposed a two-branch DL model. The model is composed of multiple CNN layers in one branch and groups of LSTM layers in another. The outputs from both branches are combined and passed through several fully connected (FC) layers, and ultimately a final sigmoid layer to predict RUL. The model performed better than deep CNN, LSTM, and multiobjective Deep Belief networks on NASA's C-MAPSS dataset. In 2021, Huang et al. \cite{huang2021novel} proposed a novel two-branch DL model that uses various features extracted from raw data. The model comprises a multilayer perception (MLP) branch working with 1D features such as RMS, Kurtosis, etc. Additionally, it employs a combination of LSTM and CNN layers in the second branch to process the 2D features generated by Wavelet transform. The model outperforms MLP, Bi-LSTM, and Multiscale-CNN on both XJTU-SY and PRONOSTIA datasets. Most recently, a two-branch DL model composed of Bi-LSTM and Bidirectional GRU (Bi-GRU) branches has been proposed by Cheng et al. in 2022 \cite{cheng2022deep}. The model achieves better results when compared to Bi-LSTM, Bi-GRU, Stacked Denoising Auto-Encoder (SDAE), Extreme Learning Machine (ELM), and MLP on the XJTU-SY dataset. Despite the numerous advantages, the existing DL models for RUL prediction of rotating machines have several drawbacks:
\begin{itemize}
    \item The multi-branch models that work directly with raw data are not effective to learn complex frequency or time-frequency features~\cite{huang2021novel}. Otherwise, models that use 1D and 2D features risk deformation or loss of information~\cite{cheng2022deep}. 
    \item The existing models often consist of basic CNN or LSTM layers, which leads to ineffectively extract feature map from the vibration signals.
    \item The performance of the current models is adversely affected by noise and anomaly data, which makes them less robust \cite{an2020data}.
\end{itemize}

To enhance the versatility of the neural network, it would be beneficial to enable it to perform an additional task alongside RUL prediction. In the datasets used for RUL prediction, such as the PRONOSTIA and XJTU datasets, bearing data is divided into multiple loads and speeds, treated as distinct operational conditions (OC). If we could utilize the neural network's potential to handle both RUL prediction and OC classification simultaneously, it would significantly contribute to the maintenance process. This capability would allow us to gain a deeper understanding of the factors that cause issues in rolling bearing machines. As a result, it will become easier to make informed maintenance decisions.

To address the above challenges, this study proposes the Robust Multi-Branch Deep Learning (Robust-MBDL) model. This model is specifically designed to predict the RUL and identify the OC of rotating machinery. The Robust-MBDL model has several advantages:

\begin{enumerate}
    \item \textbf{Feature Diversification:} Multiple types of features are utilized for RUL prediction and OC identification, including raw vibration signals, 11 time-domain features, 3 frequency-domain features (1D data), and time-frequency representation (TFR) features generated by Wavelet transformation (2D data). The use of both raw vibration data and their features improves our model's learning capacity while preserving information.   

    \item \textbf{Specialized Architecture:} Efficiently extracting various types of features requires different network architectures. This paper introduces the Robust-MBDL model, employing an advanced architecture consisting of three distinct branches: a 1D data branch, a 2D data branch, and a raw data branch. These branch architectures are largely adapted from the lightweight ResNet-34 architecture \cite{he2016deep} and the convolutional building block (CBB) \cite{cai2019effective}. They use skip connections to facilitate learning, enabling the creation of complex models with many blocks, and improving the ability to learn from complex vibration data.
    
    \item \textbf{Noise Reduction:} A noise filter was developed based on the LSTM-Autoencoder architecture to reduce noise, remove abnormal data from raw vibration signals, and thus enhance the model's robustness  \cite{zhang2019multivariate, essien2019deep, marchi2015novel}. 
    \item \textbf{Branches' fusion via Attention-based Bi-LSTM (AB-LSTM) and Global Average Pooling (GAP):} By leveraging the outputs of three data branches, the AB-LSTM and GAP algorithms enable highly accurate prediction of the RUL and precise identification of a machine's operational condition. 
\end{enumerate}

This comprehensive model architecture addresses the limitations seen in prior models by focusing on diverse feature extraction, specialized network architecture design, and noise reduction, culminating in a unified and robust framework for RUL prediction and OC identification of rotating machines.

The rest of this paper is organized as follows: Section \ref{high_level} represents the high-level architecture of our proposed robust-MBDL model. We then comprehensively present all the main components of our proposed model in Sections \ref{denoising_LSTM_Autoencode}, \ref{sec_HI}, \ref{sec_Feature_extraction}, and \ref{sec_multi_branch}. Sections \ref{experiments} and \ref{results} show our experimental setting and results. Finally, some conclusions drawn from this work are presented in Section \ref{conclusion}.  

\section{The high-level architecture of the Robust-MBDL model}\label{high_level}

\begin{figure}[h]
    \centering
    \hspace*{-1.5cm}\includegraphics[width=15cm]{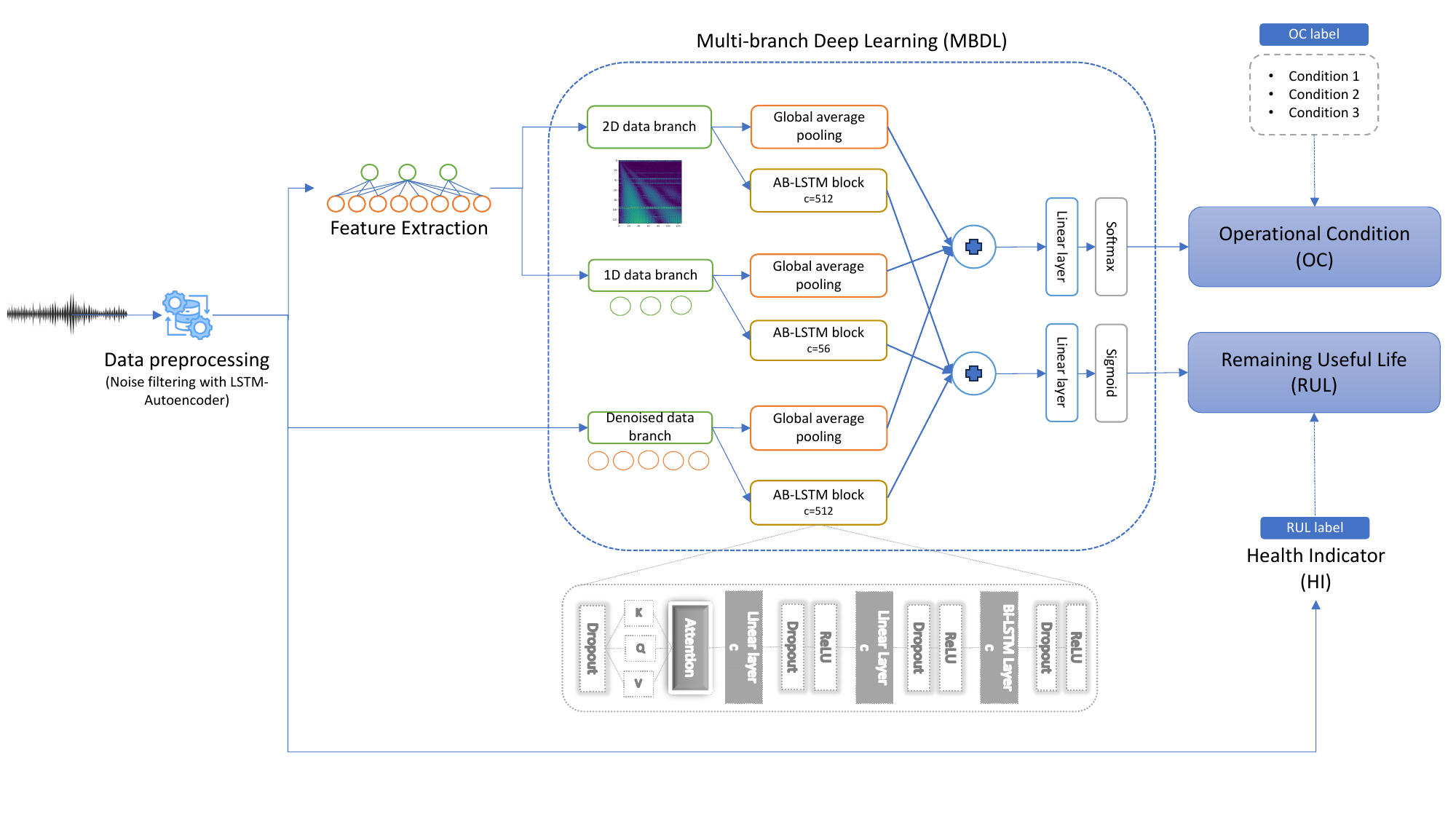}
    \caption{The high-level architecture of our proposed Robust-MBDL model}
    \label{fig:MBNNet_approach}
\end{figure}
The architecture of our proposed Robust-MBDL consists of four primary components, as shown in Figure \ref{fig:MBNNet_approach}.
\begin{itemize}
  \item Noise filtering using LSTM-Autoencoder 
  \item Feature extraction 
  \item Health Indicators (HI) construction
  \item Multi-branch deep learning (MBDL) network. 
\end{itemize}

The process starts with data denoising and abnormal data clearing through an LSTM-Autoencoder-based filter.  The denoised data are then used to extract the different features and also to construct the HI. Given the denoised data, 14 distinct 1D features (e.g., Root Mean Square, Variance, etc.) and a 2D feature are extracted (i.e. 2D feature is the spectrogram obtained via the wavelet transform). 
The MBDL network is composed of three separate branches that extract information from denoised data, 1D features, and 2D features. Two blocks, AB-LSTM and GAP, follow each branch to proficiently handle the OC identification and RUL prediction simultaneously.
\section{Noise filtering using LSTM-Autoencoder}\label{denoising_LSTM_Autoencode}
LSTM, a specialized form of RNN, effectively handles short and long-term dependencies in time series predictions by maintaining memory across numerous time steps. Unlike traditional RNN, LSTM circumvents the vanishing gradient problem during training \cite{sugiyama1991review}. It employs input, forget, and output gates to manage information flow, enabling the retention of pertinent data and discarding unnecessary information. These mechanisms significantly enhance the accuracy of time series predictions. The core of an LSTM cell involves several gates regulating information flow: the input gate controls what enters the cell, the forget gate manages what's removed from memory, and the cell state is updated by balancing incoming and outgoing information, influencing the output and hidden state. Based on these reasons, LSTM is applied in the proposed LSTM-Autoencoder model.

\begin{figure}[t]
    \centering
    \includegraphics[width=10cm]{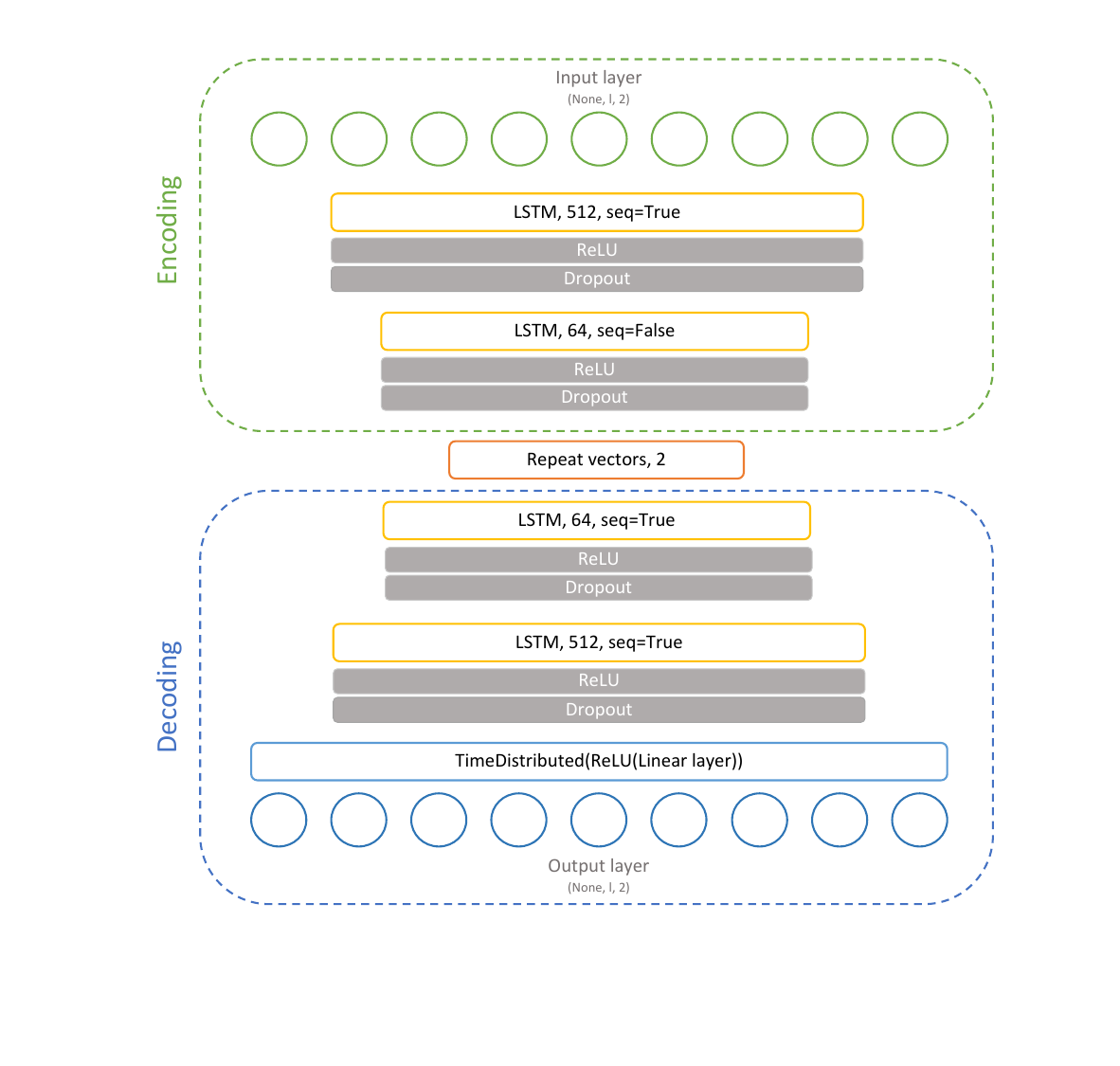}   
    \qquad
    \vspace{-1.5cm}
    \caption{The architecture of LSTM-Autoencoder}
    \label{LSTM-Autoencoder}
\end{figure}

An autoencoder is an artificial neural network widely used for learning hidden patterns of unlabeled data. An autoencoder contains two parts: an encoder and a decoder. The encoder maps the input data to hidden patterns and the decoder tries to reconstruct the output from the hidden patterns. The autoencoder is trained to minimize the difference between the input and the reconstructed output. The autoencoder has been successfully applied to different problems such as dimension reduction, anomaly detection, noise reduction, etc. Notably, both encoder and decoder in an autoencoder are designed to adapt the data types for better learning \cite{vincent2010stacked}. In our paper, the proposed autoencoder is used to reduce the noise in vibration data. To this end, the encoder and decoder are composed of LSTM layers recently mentioned to explore the short and long-term dependencies of the vibration data. The detailed structure of our LSTM-Autoencoder is presented in Fig.~\ref{LSTM-Autoencoder}.          

For more detail, the architecture contains two LSTM layers with 64 and 512 cells. To enhance model robustness, ReLU activation, and dropout layers are added after each LSTM layer, inspired by Kunang et al. \cite{kunang2018automatic}. Moreover, a repeat vector layer is employed to duplicate the previous vector. Finally, a time-distributed layer is applied to each temporal slice of the input data. During the training process, the following mean squared error (MSE) is minimized \cite{song2019combined}.
\begin{equation}
    L_{Autoencoder} = \sum^{T}_{t=1}[f_{Autoencoder}(x(t)) - x(t)]^{2},
\end{equation}
where $T$ represents the total number of segments within the training data. $f_{\text{Autoencoder}}(x(t))$ denotes the LSTM-Autoencoder output derived from the input $x(t)$ at time $t$.

The optimization process involves minimizing $L_{\text{Autoencoder}}$ via Adam Optimization \cite{kingma2014adam}. This proposed LSTM-Autoencoder is crucial for denoising vibration signals, strengthening the overall Robust-MBDL model towards higher resilience.
\section{Health Indicator (HI) construction}\label{sec_HI}
The purpose of this step is to determine the Remaining Useful Life (RUL) at every time step. We employ two popular methods for this purpose: HI construction based on the first prediction time (HI-FPT), which is inspired by the work of Huang et al. (2021) \cite{huang2021novel}, and HI construction based on Principal Component Analysis (PCA) using the Euclidean distance metric (HI-PCA), as explained in detail in Xu et al. (2022) \cite{xu2022sacgnet}.
\subsection{HI-FPT}\label{sub_sec_HIFPT}
Most industrial equipment, including rotating machines, tend to degrade only after some time of operation. Trying to predict their remaining useful life (RUL) before any signs of degradation is unreliable and unnecessary. Hence, it is crucial to detect the initial degradation time, also known as the ``First Prediction Time" (FPT) point. This time is significant because it marks the point at which the RUL prediction becomes reasonable. In this paper, the 3$\sigma$ method, which has been recognized as a simple but efficient method to detect the FPT point according to the literature \cite{lei2018machinery,li2015improved}, is applied. This method comprises two phases, which are explained below:
 \begin{itemize}
 \item Learning phase: we first select the data in the period in which the degradation does not exist, denoted $(1,T_0)$. The mean $\mu$ and the standard deviation $\sigma$ are calculated from the selected data as follows:
 \begin{equation}
 \mu=\frac{1}{T_0}\sum_{i=1}^{T_0} x_i \; \text{and} \; \sigma=\sqrt{\frac{1}{T_0}\sum_{i=1}^{T_0} (x_i-\mu)^2}
 \end{equation}
where $x_i$ represents the $i^{\text{th}}$ data point.
 \item Detecting phase: if there exist two consecutive data points that are out of the normal interval $[\mu-3\sigma, \mu+3\sigma]$, the second point is considered as the FPT point. The condition of two consecutive points is used to reduce the likelihood of making a wrong decision due to noise.
 \end{itemize}
The RUL is a function that increases linearly over time. Its maximal value is equal to $1$ at the FPT point and decreases to $0$ at the failure time, denoted by $t_{N}$. The value of RUL at an instant $ t \in [FPT, t_N]$ is calculated as follows:
\begin{equation}
RUL_t=\frac{t_{N} - t}{t_{N}-FPT}.
\end{equation}
\subsection{HI-PCA}
According to HI-PCA method, the RUL values are determined based on the covariance matrix $V$ calculated by PCA  \cite{xu2022sacgnet}. This matrix displays the shared features between time series data and its neighboring points, which accurately reflect the surrounding points' degradation trend. The calculation of the RUL value at $t^{th}$ time involves determining the average Euclidean distance from that point in $V$ to its sequential neighboring points.
\begin{equation}
RUL_{t} = \frac{1}{2}(\sqrt{\sum^{k}_{j=1}(V_{{j}} - V_{(t+1)_{j}})^{2}} + \sqrt{\sum^{k}_{j=1}(V_{{j}} - V_{(t-1)_{j}})^{2}})
\end{equation}
where $k$ represents the $k^{\text{th}}$ principal component.

\section{Feature extraction}\label{sec_Feature_extraction}
Vibration signals are initially obtained as a series of digital values representing proximity, velocity, or acceleration in the time domain. Feature extraction helps to increase the signal-to-noise ratio and underline certain patterns in vibration signals to assist the machine fault detection and prediction. In this paper, all three categories of features, including time domain, frequency domain, and time-frequency domain, are considered.
\subsection{Time-domain features}
11 popular time-domain features, including Root Mean Square, Variance, Kurtosis, etc., are used and reported in Table \ref{table_time_domain_features}. These features have proved useful in detecting machinery faults. They are simple and can be quickly calculated. However, it is difficult to detect the change in frequencies based on these features. 
\begin{table}[t]
\caption{\uppercase{Time-domain features}}
\centering
\scalebox{0.8}{
\begin{tabular}{|c|c|c|}
\hline  
No. & Formula   &  Features\\
\hline 
\multirow{2}{*}{1}    &  \multirow{2}{*}{$RMS = \sqrt{\frac{1}{n}\sum^{n}_{i=1}x^{2}_{i}}$} & \multirow{2}{*}{Root Mean Square}\\
& & \\
\multirow{2}{*}{2}    & \multirow{2}{*}{$Var = \frac{1}{n}\sum^{n}_{i=1}(x_{i}-\Bar{x})^{2}$} & \multirow{2}{*}{Variance}\\
& & \\
\multirow{2}{*}{3}    & \multirow{2}{*}{$PvT = \max( |x_{i}|)$}  & \multirow{2}{*}{Peak value} \\
& & \\
\multirow{2}{*}{4}    &  \multirow{2}{*}{$cf = \frac{PvT}{RMS}$} & \multirow{2}{*}{Crest factor}\\
& & \\
\multirow{2}{*}{5}    & \multirow{3}{*}{$Kur = \sum^{n}_{i=1}\frac{(x_{i}-\bar{x})}{n\cdot \text{var}^{2}}$}  & \multirow{2}{*}{Kurtosis}\\
& & \\
& & \\
\multirow{2}{*}{6}    &  \multirow{2}{*}{$Clf = \frac{PvT}{\frac{1}{n}\sum^{n}_{i=1} | x_{i} |}$} & \multirow{2}{*}{Clearance factor}\\
& & \\
& & \\
\multirow{2}{*}{7}     &  \multirow{2}{*}{$SF = \frac{RMS}{\frac{1}{n}\sum^{n}_{i=1} | x_{i} |}$} & \multirow{2}{*}{Shape factor}\\
& & \\
\multirow{2}{*}{8}     &  \multirow{2}{*}{$LI = \sum^{n}_{i=0} | x_{i+1} - x_{i} |$} & \multirow{2}{*}{Line integral}\\
& & \\
\multirow{2}{*}{9}     &  \multirow{2}{*}{$PP = \max(x_{i}) - \min(x_{i})$} & \multirow{2}{*}{Peak to peak value}\\
& & \\
\multirow{2}{*}{10}    &  \multirow{2}{*}{$Sk = \frac{\frac{1}{n}\sum^{n}_{i=1}(x_{i}-\bar{x})^{3}}{(\sqrt{\frac{1}{n}\sum^{n}_{i=1}(x_{i}-\bar{x})^{2}})^{3}}$} & \multirow{2}{*}{Skewness}\\
& & \\
& & \\
\multirow{2}{*}{11}    &  \multirow{2}{*}{$ IF = \frac{PvT}{\frac{1}{n}\sum^{n}_{i=1}|x_{i}|}$} & \multirow{2}{*}{Impulse factor}\\
& & \\
\hline 
\end{tabular}}
\label{table_time_domain_features}
\end{table}

\subsection{Frequency-domain features}
In reality, many types of bearing defects, such as outer race, inner race, or ball defect, can be efficiently detected in the frequency domain with the Fast Fourier Transform (FFT)\cite{nussbaumer1982fast}. We first used the FFT to convert the original signals to frequency-domain data.
\begin{equation}
X_k=\sum_{j=0}^{n-1}x_j\cdot e^{-i 2\pi k j/n}
\end{equation}
where $x_j$ and $X_k$ are the raw and frequency data, respectively.

The FFT transformation results are used to compute three frequency-domain features: FFT peak-to-peak values, energy, and power spectral density. These features are listed in Table \ref{table_frequency_domain_feature}. The features are a useful tool for stationary periodic signals but less effective for non-stationary signals that arise from time-dependent events, such as motor startup or changes in operating conditions.
\begin{table}[t]
\caption{\uppercase{Frequency-domain features}}
\centering

\scalebox{0.8}{\begin{tabular}{|c|c|c|}
\hline  
No. & Formula   &  Features\\
\hline 
\multirow{2}{*}{1}    &  \multirow{2}{*}{$r_{k} = \sum^{\infty}_{i=-\infty}x(t)e^{-iwt}$} & \multirow{2}{*}{Peak to peak value of FFT}\\
& & \\
& $PvF = \max(r_{k})$ & \\
\multirow{2}{*}{2}    & \multirow{2}{*}{$En = \sum^{N}_{k = 1} r_{k}$}  & \multirow{2}{*}{Energy of FFT}\\
& & \\
\multirow{2}{*}{3}    & \multirow{2}{*}{$PSD = \sum^{\infty}_{k = -\infty} r_{k} e^{-iwk}$}  & \multirow{2}{*}{Power spectral density of FFT} \\
& & \\
\hline 
\end{tabular}}
\label{table_frequency_domain_feature}
\end{table}
\subsection{Time-frequency domain features}
To capture the changes in frequencies over time due to the dynamic operation of rotating machines, the time-frequency features are extracted by using the Wavelet Continuous Transform (CWT)~\cite{huang2021novel,tinoco2019characterized, yoo2018novel}. The CWT uses a series of wavelets (small waves). The wavelet transform of a continuous signal $x(t)$ is defined as
 \begin{equation}
    CWT(a, b) = \frac{1}{\sqrt{c_{\psi}|a|}}\int_{-\infty}^{\infty}x(t)\psi\Big{(}\frac{t-b}{a}\Big{)}dt
\end{equation}
where $a$ in $\mathbb{R}$ and $b$ in $\mathbb{R}^{+}$ are the location parameter and the
scaling (dilation) parameter of the wavelet, respectively. $\psi(t)$ is the mother wavelet function, which is defined according to the signal inputs. In the paper, the Morlet wavelet \cite{lin2000feature} was chosen. This mother wavelet is similar to human perception (both hearing and vision). The formula for the Morlet wavelet is as follows:
\begin{equation}
    \psi(t) = e^{-\frac{\beta t^{2}}{2}}e^{j\omega_0 t}
\end{equation}
where $\beta=\omega_0^2$ and  $c_{\psi}=\sqrt{\pi/\beta}$.

It is important to mention that while feature extraction can help in predicting the RUL by highlighting key patterns in the data, it can also result in the loss or distortion of information. Therefore, in addition to the 1D and 2D features, we also incorporate denoised data as the third input for our DL model.
\section{Multi-branch Deep Learning Network}\label{sec_multi_branch}
Each type of feature recently mentioned has its own characteristics and therefore requires a specific learning mechanism. Therefore, the proposed MBDL model comprises three individual learning branches that are designed to be compatible with each type of feature.
\subsection{1D data branch}
\begin{figure}[h]
    \centering    {{\includegraphics[width=8cm]{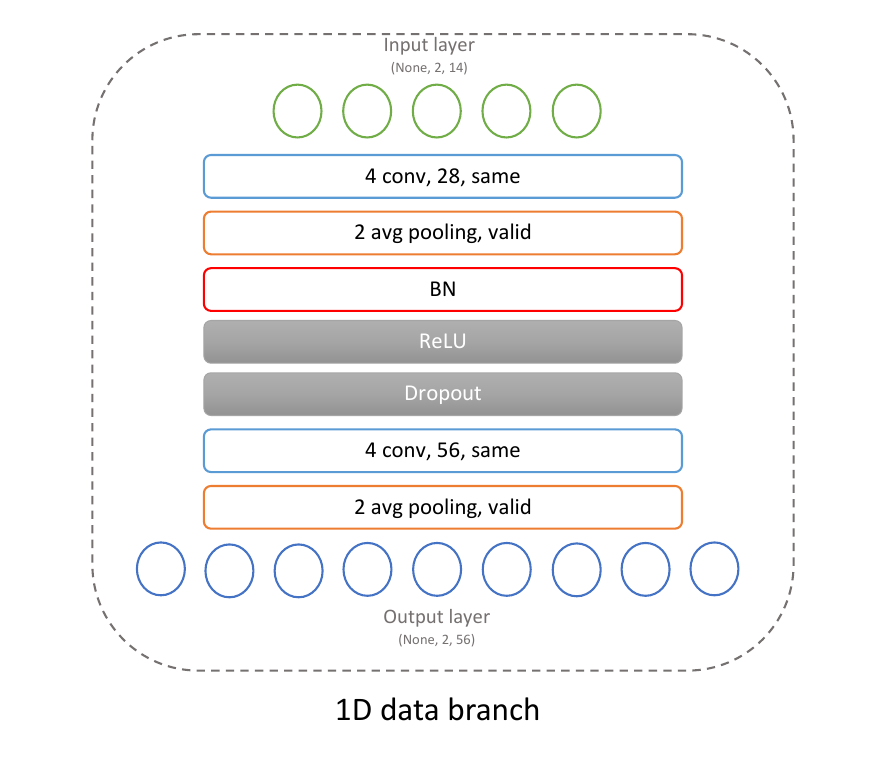}}}
    \qquad
    \vspace{-0.4cm}
    \caption{The architecture of the 1D data branch}
    \label{1D_branch_structure}
\end{figure}
This section is specifically tailored to explore the 1D data. To address this, we have empirically developed a CNN-based architecture, illustrated in Fig.~\ref{1D_branch_structure}.

The main elements of this branch consist of convolutional layers, pooling layers, batch normalization (BN), dropout layers, and the Rectified Linear Unit (ReLU) activation function layers. The convolutional layers perform operations that involve the dot product or element-wise product between an input region, defined by a sliding window, and a trainable kernel to extract pertinent information from the input data. This process generates a feature map that encapsulates essential features from the entire input dataset. The ReLU activation function, represented as $ReLU(x)=\max(0,x)$, introduces non-linear characteristics into the network. Moreover, a batch normalization block is incorporated to optimize the training process by reducing internal covariance shift and normalizing the inputs between batches \cite{kiranyaz20211d}. The pooling layers are integrated to decrease the dimensionality of the feature map by reducing redundant information. Similar to the convolutional layers, a sliding window traverses the feature map, and the average value (AVG pooling) within this window is computed. This reduction in dimensionality aims to retain essential information while improving computational efficiency.

It is important to note that the output dimension is larger than the input dimension. The purpose of this extension is to provide a more detailed and comprehensive depiction of the input information. By expanding the available space, the model becomes capable of capturing more intricate and meaningful relationships between the features, which ultimately improves its ability to learn from the data. 
\subsection{2D data branch}
\begin{figure}[H]
    \centering    
    \includegraphics[width=24cm]{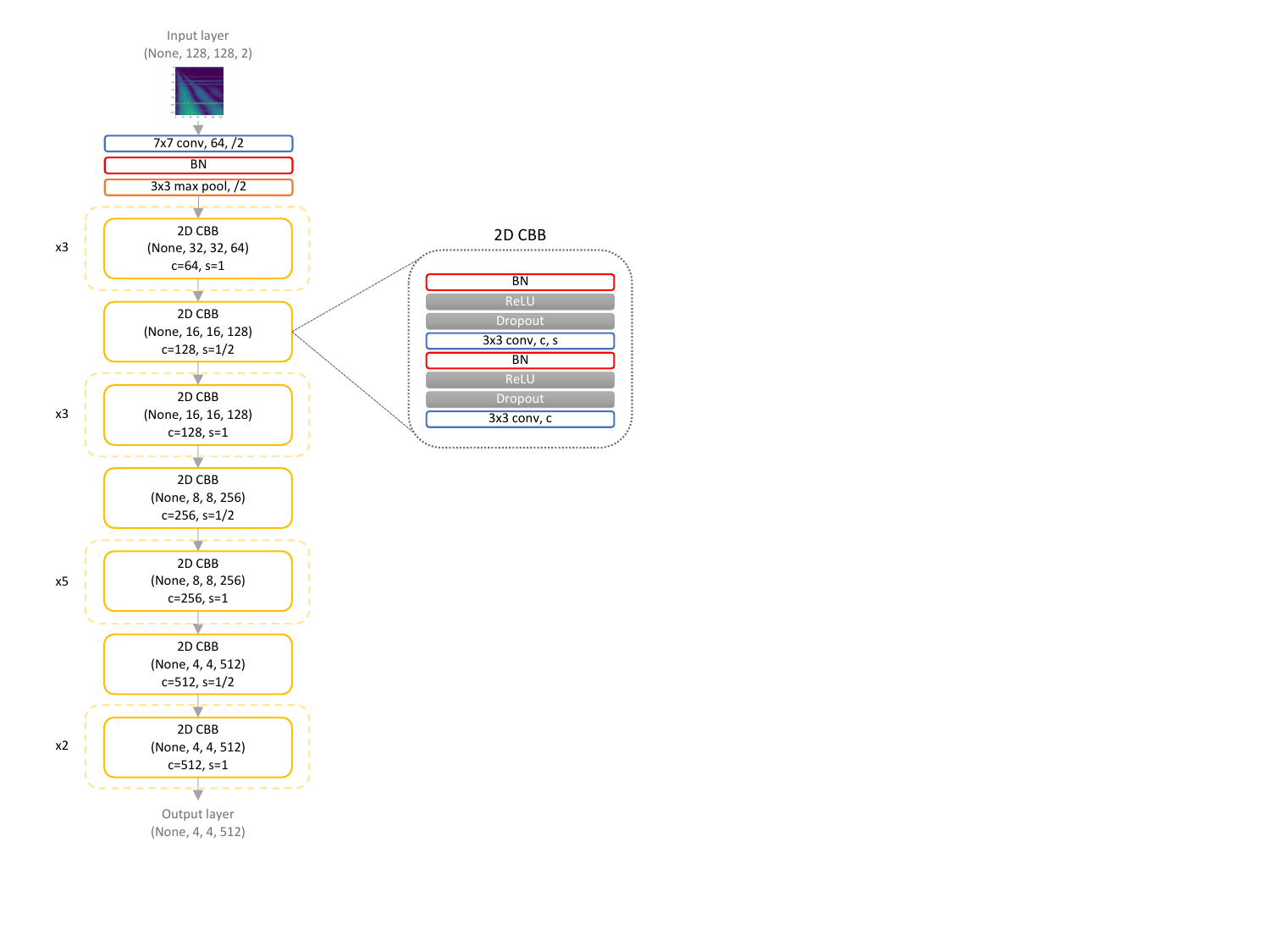}
    \qquad
    \vspace{-2cm}
    \caption{The architecture of the 2D data branch}
    \label{fig:modified_resnet34}
\end{figure}

This branch, as shown in Fig.~\ref{fig:modified_resnet34} is designed to process the 2D feature (time-frequency domain features) obtained from the wavelet transform. The underlying structure of this branch relies on ResNet-34 ~\cite{he2016deep}. The ResNet-34 is a lightweight yet effective deep learning architecture with 34 layers that utilizes residual blocks. It integrates shortcuts and skip connections, facilitating the training of remarkably deep networks and mitigating the complexities associated with identifying intricate features within data. In addition, recognizing the limitations of traditional residual blocks in handling complex vibration data with sudden changes, we propose replacing them with the convolutional building block (CBB), proposed by Shaofeng Cai et al. in 2019 \cite{cai2019effective}. For more details, our 2D feature branch consists of four groups of CBBs. Each group contains 3, 3, 5, and 2 CBBs, respectively. Finally, in each CBB, we employ batch normalization (BN), ReLU activation, and a dropout layer with a dropout rate of 0.2.
\subsection{Denoised data branch}
\begin{figure}[H]
    \includegraphics[width=30cm]{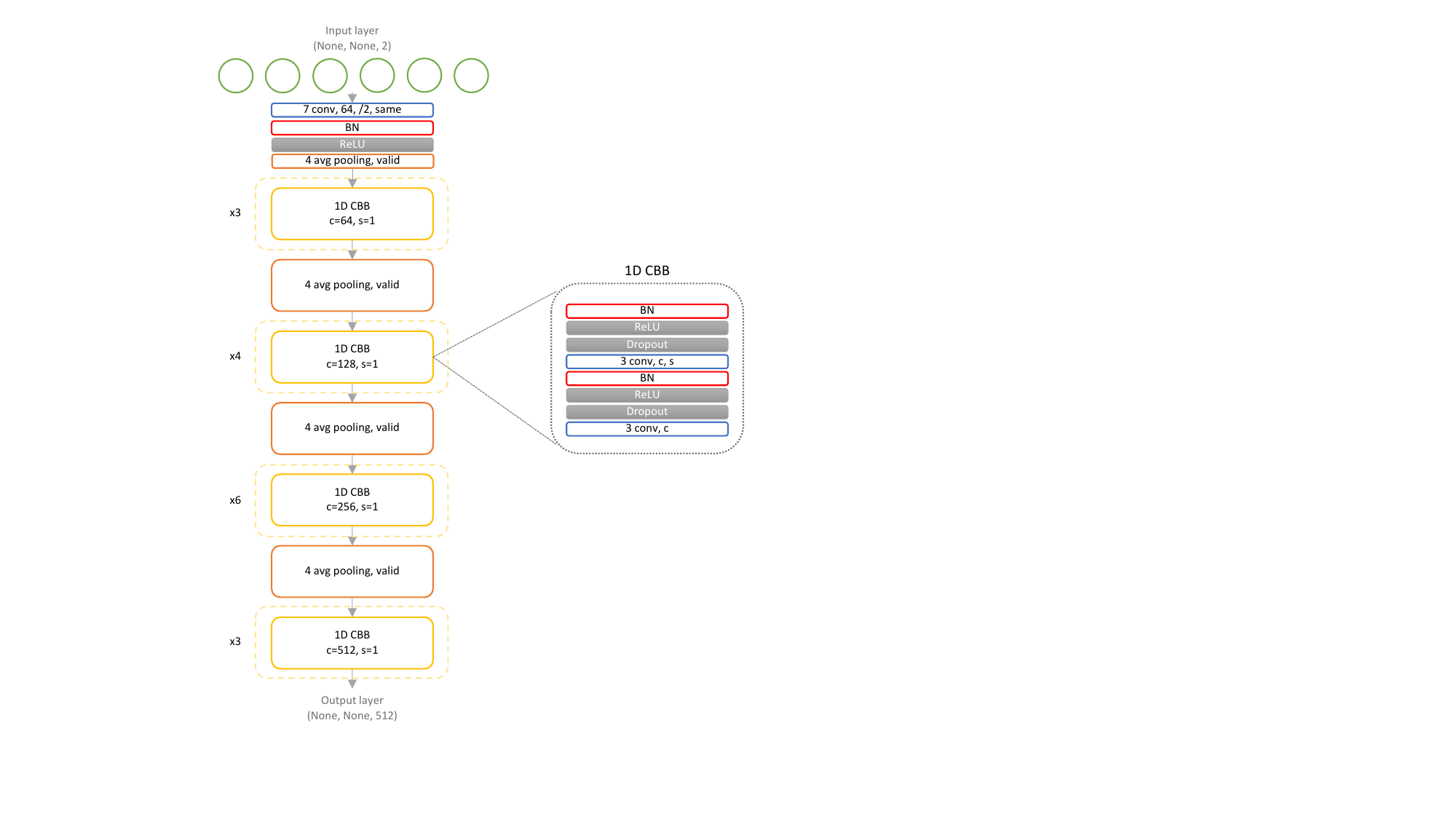}
    \qquad
    \vspace{-2cm}
    \caption{The architecture of the denoised data branch}
    \label{Architecture_denoised_data_branch}
\end{figure}
The purpose of this branch is to analyze the vibration data that is directly obtained from the denoising LSTM-Autoencoder. The direct explosion of the denoised vibration data is important since the information may be lost or deformed during the extraction of 1D and 2D data. The architecture of this branch (see Fig.~\ref{Architecture_denoised_data_branch}) was designed as an extension of the 2D feature branch, specifically tailored to better explore the vibration features. In particular, this branch consists of the same number of CBBs as that of our 2D feature branch; however, 1D convolutional layers were used instead of the 2D convolutional layers. In addition, an average pooling layer with a window size of 4 was added after each CBB to capture all relevant features by considering their relationship, while the overall shape is smaller.
\subsection{AB-LSTM and GAP}\label{sec_ab_lstm}
The AB-LSTM blocks are designed based on the Bi-LSTM architecture to optimize the RUL prediction task. The Bi-LSTM integrates both forward and backward hidden layers. This design allows the model to assimilate information from both past and future sequences, proving superior in tasks like RUL prediction compared to traditional LSTM networks \cite{jin2022bi}. Furthermore, self-attention mechanisms are also used to assist the Bi-LSTM in identifying significant input segments, leading to quicker convergence and improving the model performance \cite{chien2023accurate, zhou2023attention}. For more details, Vaswani et al. \cite{vaswani2017attention} describe attention mechanisms as ``mapping a query and a set of key-value pairs to an output, where the query, keys, values, and output are all vectors. The output is computed as a weighted sum of the values, where the weight assigned to each value is computed by a compatibility function of the query with the corresponding key''. Let $Q, K, V$ denote the query, key, and value vectors, respectively. The attention mechanism is described mathematically as follows:
\begin{equation}
    Attention(Q, K, V) \\
    = Softmax[\frac{Q K^{\top}}{\sqrt{d_k}}] V
\end{equation}
and each head
\begin{equation}
    H_{i} = Attention(QW^{Q}_{i}, KW^{K}_{i}, VW^{V}_{i})
\end{equation}
where $W^{Q}_{i}, W^{K}_{i} \in \mathbb{R}^{d_h \times d_v}$, $W^{V}_{i}\in \mathbb{R}^{d_h \times d_v}$ are weight matrices, and $d_v$, $d_k$ denote the projection subspaces' hidden dimensions. $\frac{1}{\sqrt{d_{k}}}$ is the scale factor that helps dot-product attention be faster when using a feed-forward network. Each $H_{i}$ is concatenated into a matrix $W^{O}\in \mathbb{R}^{hd_{v} \times d_{h}}$ that integrates with projections to compile the data gathered from various positions on particular sub-spaces.
\begin{equation}
    Attention(Q, K, V) = Concat(H_{1}, ..., H_{h})W^{O}
\end{equation}
In this paper, the number of heads (parallel attention layers) was fixed at $h = 16$ according to our tests. Hence, $\frac{d_{v}}{h}=\frac{d_{k}}{h}=32$. The overall computing cost is comparable to that of single-head attention with full dimensionality because of the lower dimension of each head. The three AB-LSTM blocks' outputs are concatenated and passed to a linear layer with a Sigmoid activation function, ensuring a final output range of (0,1) \cite{rasamoelina2020review}.

The GAP layers are designed to automatically identify the machine's OC. The GAP layers are designed to automatically identify the machine's operating characteristics. The idea behind using GAP is to calculate the average of each feature map and feed it into a softmax layer, rather than using a fully connected layer. Compared to a fully connected layer, GAP is more suited to convolutional structures as it enforces correspondences between feature maps and categories and is more tolerant of spatial translations of the input. Additionally, there are no parameters to optimize \cite{lin2013network}.  Finally, three GAP layers' outputs are concatenated and fed to a linear layer with softmax activation to compute OC probabilities.
\section{EXPERIMENTAL SETTINGS}\label{experiments}

\subsection{Datasets}
In this paper, our proposed model was evaluated using the two benchmark datasets: XJTU-SY~\cite{wang2018hybrid} and PRONOSTIA~\cite{nectoux2012pronostia}. 
\begin{table}[H]
\caption{\uppercase{The XJTU-SY bearing dataset}~\cite{wang2018hybrid}}
\centering
\resizebox{\textwidth}{!}{
\begin{tabular}{|c|c|c|c|c|}
\hline  
OC  & Bearing dataset & Bearing lifetime ($t_{N}$) & Estimated FPT & Real FPT\\
\hline 
Condition 1 (2100 rpm, 12000 N) & $Bearing1_{-}1$  & 2 h 3 m & 1 h 16 m & -\\
 & $Bearing1_{-}2$  & 2 h 3 m & 44 m &  -\\
 & $Bearing1_{-}3$  & 2 h 38 m & 1 h & -\\
 & $Bearing1_{-}4$  & 2 h 38 m & 1 h 20 m  & -\\
 & $Bearing1_{-}5$  & 52 m & 39 m &  -\\
\hline 
Condition 2 (2250 rpm, 11000 N) & $Bearing2_{-}1$  & 8 h 11 m & 7 h 35 m & -\\
 & $Bearing2_{-}2$  & 2 h 41 m & 48 m & -\\
 & $Bearing2_{-}3$  & 8 h 53 m & 5 h 27 m & -\\
 & $Bearing2_{-}4$  & 42 m & 32 m & -\\
 & $Bearing2_{-}5$  & 5 h 39 m & 2 h 21 m & -\\
\hline 
Condition 3 (2400 rpm, 10000 N) & $Bearing3_{-}1$  & 42 h 18 min & 39 h 4 min & -\\
 & $Bearing3_{-}2$  & 41 h 36 m & 20 h 30 m & -\\
 & $Bearing3_{-}3$  & 6 h 11 m  & 5 h 40 m& -\\
 & $Bearing3_{-}4$  & 25 h 15 m & 23 h 38 m & -\\
 & $Bearing3_{-}5$  & 1 h 54 m  & 9 m & -\\
 \hline 
\end{tabular}
}
\label{data_XJTU}
\end{table}

The XJTU-SY dataset was created by the Institute of Design Science and Basic Component at Xi’an Jiaotong University. It consists of 15 trials under three different operational conditions, referred to as from $Bearing1_{-}1$ to $Bearing3_{-}5$ in Table~\ref{data_XJTU}. The vibraration data was collected from two PCB 352C33 accelerometers, each of which was installed at a 90° angle, with one on the horizontal axis and the other on the vertical axis, to collect data. Each data segment contains 32768 data points and was collected in one minute.

The PRONOSTIA dataset was published by the FEMTO-ST Institute in France and used in the 2012 IEEE Prognostic Challenge~\cite{nectoux2012pronostia}. 
It consists of 17 accelerated run-to-failures on a small-bearing test rig, referred to as from $Bearing1_{-}1$ to $Bearing3_{-}3$ (Table~\ref{table1}). The bearing was operated under three operating conditions with different levels of rotation speed and load.
The vibration signals include vertical and horizontal data, which were gathered by two miniature accelerometers positioned at $90^{\circ}$. Each data segment contains 2560 data points and was collected in 0.1 seconds.

\begin{table}[H]
\caption{\uppercase{The PRONOSTIA bearing dataset}~\cite{nectoux2012pronostia}}
\centering
\resizebox{\textwidth}{!}{
\begin{tabular}{|c|c|c|c|c|}
\hline  
OC  & Bearing dataset & Bearing lifetime $(t_{N})$ & Estimated FPT & Real FPT\\
\hline 
Condition 1 (1800 rpm, 4000 N) & $Bearing1_{-}1$  & 28030 s & 5000 s & -\\
 & $Bearing1_{-}2$  & 8710 s & 660 s & -\\
 & $Bearing1_{-}3$  & 18020 s & 5740 s & 5730 s
\\
 & $Bearing1_{-}4$  & 11390 s & 340 s & 339 s\\
 & $Bearing1_{-}5$  & 23020 s & 1600 s & 1610 s
\\
 & $Bearing1_{-}6$  & 23020 s & 1460 s & 1460 s
\\
 & $Bearing1_{-}7$  & 15020 s & 7560 s & 7570 s\\
\hline 
Condition 2 (1650 rpm, 4200 N) & $Bearing2_{-}1$  & 9110 s & 320 s & -\\
 & $Bearing2_{-}2$  & 7970 s & 2490 s & -\\
 & $Bearing2_{-}3$  & 12020 s & 7530 s & 7530 s
\\
 & $Bearing2_{-}4$  & 6120 s & 1380 s & 1390 s
\\
 & $Bearing2_{-}5$  & 20020 s & 3100 s & 3090 s
\\
 & $Bearing2_{-}6$  & 5720 s & 1280 s & 1290 s
\\
 & $Bearing2_{-}7$  & 1720 s & 580 s & 580 s
\\
\hline 
Condition 3 (1500 rpm, 5000 N) & $Bearing3_{-}1$  & 5150 s & 670 s & -\\
 & $Bearing3_{-}2$  & 16370 s & 1330 s & -\\
 & $Bearing3_{-}3$  & 3520 s  & 800 s & 820 s
\\
 \hline 
\end{tabular}
}
\label{table1}
\end{table}
Tables~\ref{data_XJTU} and~\ref{table1} show detailed information on the two datasets. $h$, $m$, and $s$ denote hours, minutes, and seconds, respectively. The tables report the estimated and real FPT. The estimated FPT is calculated using the FPT detection method in subsection \ref{sub_sec_HIFPT}, and the real FPT is taken from the dataset if available.
\subsection{Data splitting}
As almost all the state-of-the-art systems proposed for RUL detection on the XJTU-SY and PRONOSSTIA datasets used the data splitting methods from \cite{huang2021novel} and~\cite{xu2022sacgnet}, respectively.
Therefore, we obey these data-splitting methods from these papers to compare our experimental results to state-of-the-art systems.
In particular, two splitting methods are proposed and referred to as the operating condition-dependent rule (OC-dependent rule) and the operating condition-independent rule (OC-independent rule), respectively.
\begin{itemize}
\item OC-independent method: This data-splitting method does not consider the operating condition of bearings \cite{huang2021novel}. For a specific test, one bearing is randomly chosen as the evaluating data, and all the other bearings in the dataset are considered the training data regardless of the bearings' operating conditions.
\item OC-dependent method: The data-splitting method takes into account the bearing's operating condition \cite{xu2022sacgnet}. Within each OC, two bearings are assigned to be the training data, while the remaining bearings are reserved for model evaluation. 
\end{itemize}
\subsection{Validation methods}
To evaluate the performance of our model in RUL forecasting, we calculate the root mean square error (RMSE) and the mean absolute error (MAE) using the following equations:
\begin{equation}
    RMSE = \sqrt{\sum_{t=FPT}^{t_{N}}\frac{({RUL}_{t} - \widehat{RUL}_t)^{2}}{t_{N}-FPT}}
\end{equation}
\begin{equation}
    MAE = \sum_{t=FPT}^{t_{N}}\frac{|RUL_t - \widehat{RUL}_t|}{t_{N}-FPT}
\end{equation}

The accuracy of the model in OC identification is determined by the accuracy score (Acc).
\begin{equation}
\text{Acc} =   \frac{M}{P}\times 100 
\end{equation}
where $M$ denotes the number of well-classified segments among $P$ classified segments. 
\subsection{Loss Functions} 
We used the mean squared logarithmic error (MSLE) \cite{rengasamy2020asymmetric} to calculate the difference between the real RUL ($RUL_t$) and the RUL estimated by our Robust-MBDL model ($\widehat{RUL}_t$) during both the training and testing phases:
\begin{equation}
    L_{RUL} = \sum^{t_{N}}_{t=FPT} \frac{[log(RUL_t + 1) - log(\widehat{RUL}_t + 1)]^{2}}{t_{N}-FPT}
\end{equation}
It is noted that in the above equation, the RUL values are increased by 1 to prevent taking the logarithm of zero when the RUL equals $0$.

For the OC classification task, We employed categorical cross-entropy loss \cite{zhang2018generalized}, a widely used loss function for multi-class classification problems \cite{zhang2022hybrid}. Let $m$ denote the total number of possible operational conditions; $OC=(c_1,c_2,...,c_m)$ represents the real operational condition; $\widehat{OC}=(\hat{c}_1,\hat{c}_2,...,\hat{c}_m)$ represents the operational condition classified by our model. The cross-entropy loss can be calculated as
\begin{equation}
    L_{OC} = -\sum^{m}_{i=1}c_i\cdot \log(\hat{c}_i)
\end{equation}

Our model simultaneously addresses RUL prediction and OC classification. The two above loss functions are then combined to form the following global loss function: 
\begin{equation}
    L = \lambda L_{OC} + (1-\lambda) L_{RUL}
\end{equation}
where $\lambda$ is a real number that ranges between 0 and 1. By adjusting the value of $\lambda$, two things can be achieved: \textit{(i)} offset any imbalances between the two loss functions in the global one, and \textit{(ii)} give varying degrees of importance to each task depending on the particular study case. In our paper, we determined through experimentation that $\lambda$ is set to 0.6. 
\subsection{Deep Neural Network Implementation}
In this study, we implemented all proposed deep neural networks using the Tensorflow framework and utilized the Root Mean Squared Propagation (RMSProp) method for model optimization \cite{li2022physics, ruder2016overview}. We conducted all experiments on an Nvidia A100 GPU.

\begin{table}[H]
\caption{\uppercase{Parameters of training process.}}
\centering
\scalebox{0.9}{
\begin{tabular}{|c|c|c|c|c|}
\hline  
Model & Optimizer & Learning rate & batch size & epochs \\
\hline 
MBDL & RMSProp  & 1e-4  & 16  & 1000\\
LSTM-Autoencoder & RMSProp & 1e-4 & 16 & 300\\
\hline 
\end{tabular}
\label{learning_parameters}
}
\end{table}
Table~\ref{learning_parameters} details the specific settings applied during the training processes for both the denoised LSTM-Autoencoder and the MBDL parts. Moreover, it is crucial to optimize the number of attention heads as it greatly impacts the model's performance \cite{povey2018time}. Table \ref{head_numbers} shows results for different numbers of heads tested. 16 attention heads were selected to enhance RUL predictions by allowing the model to focus on critical input aspects.
\begin{table}[H]
\caption{\uppercase{Model's performance with respect to the different head sizes.}}
\centering
\scalebox{0.9}{
\begin{tabular}{|c|ccc|}
\hline  
  Number of heads     & OC Acc & MAE        & RMSE   \\
\hline 
32   & 20.9446       & 0.2104 & 0.2653\\
24   & 27.6873  & 0.2319 &  0.286\\
16   & 37.8936  & 0.206 &  0.2566\\
8    & 30.4622  & 0.2203 &  0.2857\\
\hline 
\end{tabular}
}
\label{head_numbers}
\end{table}
\section{Experimental Results and Discussions}\label{results}
We evaluated the performance of our proposed Robust-MBDL model for various scenarios, including RUL prediction and OC identification, using the PRONOSTIA and XJTU-SY datasets, with both OC-dependent and OC-independent rules, with and without the denoised LSTM-Autoencoder. The model's performance was also compared to various state-of-the-art ones, including BLSTM~\cite{huang2019bidirectional}, MLP and DCNN–MLP~\cite{huang2021novel}, SACGNet~\cite{xu2022sacgnet}, and MSCNN~\cite{zhu2018isolation}. The obtained results are reported in Tables \ref{XJTU_independent}, \ref{XJTU_dependent}, \ref{PHM_independent}, and \ref{Pronostia_dependent}.

\begin{table}[H]
\caption{\uppercase{Results of the performance analysis for the XJTU-SY dataset with OC-independent rule.}}
\centering
\resizebox{\textwidth}{!}{
\begin{tabular}{|c|cc|cc|cc|cc|ccc|ccc|}
\hline  
  \textbf{Test bearing}   & \multicolumn{2}{c|}{\textbf{MLP}~\cite{huang2021novel}}  & \multicolumn{2}{c|}{\textbf{BLSTM}~\cite{huang2019bidirectional}}  & \multicolumn{2}{c|}{\textbf{MSCNN}~\cite{zhu2018isolation}} & \multicolumn{2}{c|}{\textbf{DCNN–MLP}~\cite{huang2021novel}} & \multicolumn{3}{c|}{\textbf{Robust-MBDL w/o denoise}} & \multicolumn{3}{c|}{\textbf{Robust-MBDL w/ denoise}}\\

    & RMSE & MAE & RMSE & MAE & RMSE & MAE & RMSE & MAE & RMSE & MAE & Acc$(\%$) & RMSE & MAE & Acc$(\%$)\\
\hline 
$Bearing1_{-}1$  & 0.274 & 0.240 & 0.228 & 0.191 & 0.242 & 0.213 & 0.206 & 0.176 & 0.0944 & 0.0745 & \textbf{100.0} & \textbf{0.0922} & \textbf{0.0739} & \textbf{100.0}\\
$Bearing1_{-}2$  & 0.313 & 0.270 & 0.305 & 0.231 & 0.262 & 0.229 & 0.240 & 0.207 & 0.0453 & 0.037 & \textbf{100.0} &  \textbf{0.033} & \textbf{0.021} &   \textbf{100.0}\\
$Bearing1_{-}3$  & 0.261 & 0.221 & 0.130 & 0.106 & 0.184 & 0.155 & 0.178 & 0.151 & \textbf{0.0552} & \textbf{0.049} & 100.0 &  0.057 & 0.52 & \textbf{100.0}\\
$Bearing1_{-}5$  & 0.318 & 0.265 & 0.362 & 0.314 & 0.215 & 0.181 & 0.184 & 0.155 & 0.0592 & 0.0531 & \textbf{100.0} &  \textbf{0.0491} & \textbf{0.0376} &  \textbf{100.0}\\
$Bearing2_{-}1$  & 0.203 & 0.172 & 0.152 & 0.129 & 0.148 & 0.126 & 0.117 & 0.099 & \textbf{0.0867} & \textbf{0.0806} & \textbf{100.0} &  0.0877 & 0.0803 & \textbf{100.0}\\
$Bearing2_{-}2$  & 0.266 & 0.214 & 0.134 & 0.094 & 0.232 & 0.194 & 0.122 & 0.102 & 0.0555 & 0.0453 & \textbf{100.0} & \textbf{0.0365} & \textbf{0.0321} & \textbf{100.0}\\
$Bearing2_{-}3$  & 0.230 & 0.204 & 0.216 & 0.170 & 0.199 & 0.164 & 0.158 & 0.126 & 0.0588 & 0.0525 & \textbf{100.0} & \textbf{0.0576} & \textbf{0.0512} & \textbf{100.0}\\
$Bearing2_{-}4$  & 0.251 & 0.213 & 0.311 & 0.267 & 0.231 & 0.195 & 0.177 & 0.141 & \textbf{0.0771} & 0.0657 & \textbf{100.0} & 0.0775 & 0.0639 & \textbf{100.0}\\
$Bearing2_{-}5$  & 0.234 & 0.202 & 0.308 & 0.278 & 0.108 & 0.090 & 0.0918 & 0.075 & 0.0596 & 0.0505 & \textbf{100.0} & \textbf{0.0429} & \textbf{0.0398} & \textbf{100.0}\\
$Bearing3_{-}1$  & 0.305 & 0.262 & 0.351 & 0.297 & 0.247 & 0.214 & 0.244 & 0.204 & 0.0575 & 0.0489 & \textbf{100.0} & \textbf{0.0509} & \textbf{0.0418} & \textbf{100.0}\\
$Bearing3_{-}3$  & 0.318 & 0.276 & 0.188 & 0.162 & 0.191 & 0.156 & 0.158 & 0.129 & 0.0575 & 0.0459 & \textbf{100.0} &  \textbf{0.0365} & \textbf{0.0214} & \textbf{100.0}\\
$Bearing3_{-}4$  & 0.252 & 0.220 & 0.175 & 0.135 & 0.165 & 0.139 & 0.132 & 0.107 & 0.0837 & 0.0709 & \textbf{100.0} &  \textbf{0.0792} & \textbf{0.0708} & \textbf{100.0}\\
$Bearing3_{-}5$  & 0.376 & 0.310 & 0.305 & 0.251 & 0.267 & 0.225 & 0.266 & 0.219 & 0.0733 & 0.0598 & \textbf{100.0} & \textbf{0.0685} & \textbf{0.0517} & \textbf{100.0}\\
\hline 
\end{tabular}
}
\label{XJTU_independent}
\end{table}
\begin{table}[H]
\caption{\uppercase{Results of the performance analysis for the PRONOSTIA dataset with OC-independent rule.}}
\centering
\resizebox{\textwidth}{!}{
\begin{tabular}{|c|cc|cc|cc|cc|cc|cc|}
\hline  
  \textbf{Test bearing}   & \multicolumn{2}{c|}{\textbf{MLP}~\cite{huang2021novel}}  & \multicolumn{2}{c|}{\textbf{BLSTM}~\cite{huang2019bidirectional}}  & \multicolumn{2}{c|}{\textbf{MSCNN}~\cite{zhu2018isolation}} & \multicolumn{2}{c|}{\textbf{DCNN–MLP}~\cite{huang2021novel}} & \multicolumn{2}{c|}{\textbf{MBDL w/o denoise}} & \multicolumn{2}{c|}{\textbf{Robust-MBDL w/ denoise}}\\

    & RMSE & MAE & RMSE & MAE & RMSE & MAE & RMSE & MAE & RMSE & MAE & RMSE & MAE \\
\hline 
$Bearing1_{-}1$  & 0.332 & 0.277 & 0.268 & 0.245 & 0.152 & 0.122  & 0.194 & 0.161 & 0.158 & 0.121 & \textbf{0.0864} & \textbf{0.0699} \\
$Bearing1_{-}2$  & 0.256 & 0.213 & 0.281 & 0.242 & 0.484 & 0.386 & 0.254 & 0.219 & 0.167 & 0.146 & \textbf{0.0964} & \textbf{0.0854} \\
$Bearing1_{-}3$  & 0.235 & 0.186 & 0.331 & 0.270 & 0.251 & 0.208 & 0.199  & 0.164 & \textbf{0.135} & 0.112 & 0.1467 & \textbf{0.0691} \\
$Bearing1_{-}4$  & 0.515 & 0.439 & 0.513 & 0.443 & 0.397 & 0.329 & 0.132  & 0.107 & 0.101 & 0.081 & \textbf{0.1038} & \textbf{0.0768} \\
$Bearing1_{-}5$  & 0.107 & 0.320 & 0.208 & 0.174 & 0.326 & 0.276 & 0.187  & 0.158 & 0.165 & 0.136 & \textbf{0.1027} & \textbf{0.0779} \\
$Bearing1_{-}6$  & 0.480 & 0.480 & 0.329 & 0.278 & 0.340 & 0.273 & 0.328 & 0.270 & 0.088 & 0.071 & \textbf{0.0746} & \textbf{0.0593} \\
$Bearing1_{-}7$  & 0.170 & 0.153 & 0.165 & 0.141 & 0.357 & 0.299 & 0.205 & 0.172 & \textbf{0.088} & \textbf{0.071} & 0.0997 & 0.0822 \\
\hline 
\end{tabular}
}
\label{PHM_independent}
\end{table}

\begin{figure}[H]
    \centering
    \begin{subfigure}[b]{0.45\linewidth}
        \centering
        \includegraphics[width=\linewidth]{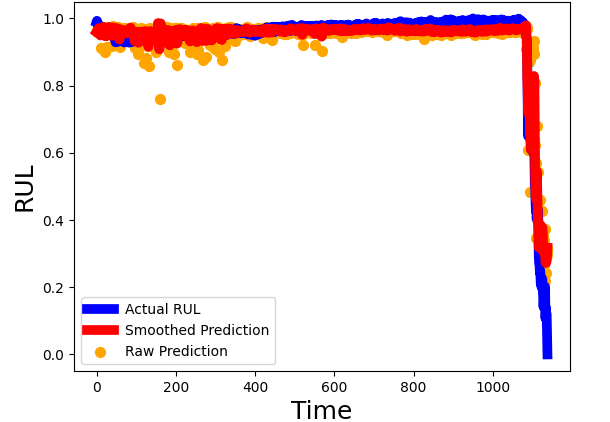}
        \caption{$Bearing1_{-}4$, PRONOSTIA dataset}
        \label{fig:hi-pca}
    \end{subfigure}
    \hfill
    \begin{subfigure}[b]{0.45\linewidth}
        \centering
        \includegraphics[width=\linewidth]{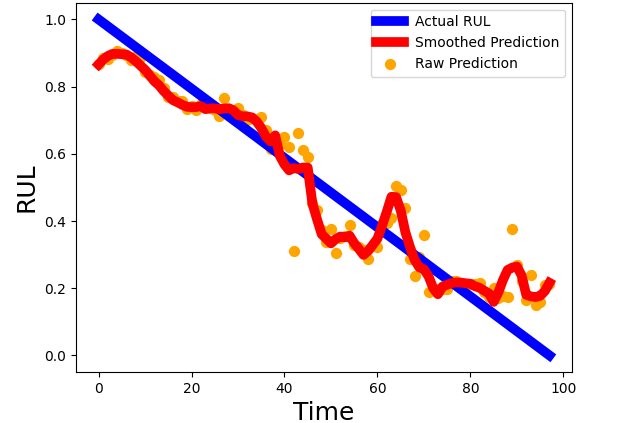}
        \caption{$Bearing1_{-}3$, XJTU-SY dataset}
        \label{fig:hi-fpt}
    \end{subfigure}
    \caption{Illustration of the RUL prediction by the Robust-MBDL model.}
    \label{fig:fig_7}
\end{figure}

The results presented in Table~\ref{XJTU_independent} and Table~\ref{PHM_independent} demonstrate the superior performance of our proposed Robust-MBDL model under the OC-independent rule for data splitting. Whether the denoised LSTM-Autoencoder is applied or not, it outperforms the state-of-the-art models for RUL prediction in terms of RMSE and MAE scores across all bearing types. Fig.~\ref{fig:fig_7} shows an example of the RUL prediction for $Bearing1_{-}3$ and $Bearing1_{-}4$. We consistently observe minimal disparity between actual and predicted RUL, providing strong evidence of our approach's reliability and effectiveness.

Regarding the OC identification task, the network shows exceptional performance, achieving 100\% accuracy for all bearing types. It is important to highlight that by training with two tasks (RUL prediction and OC classification) simultaneously, the proposed models are able to learn the complex relationships between the operating conditions of the bearings and their degradation patterns, leading to a high performance of these models. Finally, utilizing the denoised LSTM-Autoencoder, the Robust-MBDL shows outstanding performance in most bearings, proving the efficacy and necessity of the data denoising.

\begin{table}[H]
\caption{\uppercase{Results of the performance analysis for the XJTU-SY dataset with the OC-dependent rule.}}%
\centering
\resizebox{\textwidth}{!}{
\begin{tabular}{@{}|l|ll|lll|lll|@{}}
\hline
\textbf{Type}  & \multicolumn{2}{c|}{\textbf{SACGNet}~\cite{xu2022sacgnet}} & \multicolumn{3}{c|}{\textbf{Robust-MBDL w/o denoise}} & \multicolumn{3}{c|}{\textbf{Robust-MBDL w/ denoise}}\\
   & RMSE & MAE & RMSE & MAE & Acc & RMSE & MAE & Acc \\
\hline
Bearing1-3     & 0.147 & 0.117 & \textbf{0.126} & 0.076 & \textbf{100.0} & 0.139 & \textbf{0.072} & \textbf{100.0}\\
Bearing1-4     & 0.166 & 0.088 & \textbf{0.08}  & 0.043 & \textbf{4.91}  & 0.087 & \textbf{0.035} &  0.0 \\
Bearing1-5     & 0.360 & 0.206 & 0.199 & 0.093 & 98.07 & \textbf{0.177} & \textbf{0.091} & \textbf{100.0}\\
\hline
Bearing2-3     & 0.320 & 0.307 & \textbf{0.133} & \textbf{0.087} & 85.17 & 0.218 & 0.164 & \textbf{85.74}\\
Bearing2-4     & 0.511 & 0.428 & \textbf{0.105} & \textbf{0.056} & 88.09 & 0.223 & 0.103 & \textbf{90.47}\\
Bearing2-5     & 0.341 & 0.249 & \textbf{0.189} & \textbf{0.123} & 66.07 & 0.201 & 0.169 & \textbf{77.87}\\
\hline
Bearing3-3     & 0.369 & 0.256 & \textbf{0.035} & \textbf{0.018} & \textbf{99.73} & 0.177 & 0.054 & 97.8437\\
Bearing3-4     & 0.193 & 0.069 & \textbf{0.038} & \textbf{0.021} & 29.17 & 0.159 & 0.129 & \textbf{87.78}\\
Bearing3-5     & 0.500 & 0.447 & \textbf{0.263} & \textbf{0.231} & \textbf{96.49} & 0.312 & 0.24 & \textbf{96.49}\\
\hline
\end{tabular}
}
\label{XJTU_dependent}
\end{table}
\begin{table}[H]
\caption{\uppercase{Results of the performance analysis for the PRONOSTIA dataset with OC-dependent rule.}}%
\centering
\resizebox{\textwidth}{!}{
\begin{tabular}{@{}|l|ll|lll|lll|@{}}
\hline
\textbf{Type}  & \multicolumn{2}{c|}{\textbf{SACGNet}~\cite{xu2022sacgnet}} & \multicolumn{3}{c|}{\textbf{Robust-MBDL w/o denoise}} & \multicolumn{3}{c|}{\textbf{Robust-MBDL w/ denoise}}\\
   & RMSE & MAE & RMSE & MAE & Acc & RMSE & MAE & Acc \\
\hline
Bearing1-3     & 0.101 & 0.041 & 0.0624 & \textbf{0.0241} & 99.3341 & \textbf{0.0594} & 0.0281 & \textbf{99.4451}\\
Bearing1-4     & 0.230 & 0.157 & 0.045 & 0.0222 & \textbf{99.4732} & \textbf{0.0394} & \textbf{0.0213} & 97.2783\\
Bearing1-5     & \textbf{0.197} & \textbf{0.077} & 0.2407 & 0.1953 & \textbf{99.3918} & 0.2259 & 0.1777 & 99.2615\\
Bearing1-6     & 0.205 & 0.079 & 0.1376 & 0.0879 & 99.5656 & \textbf{0.1304} & \textbf{0.079} & \textbf{99.305}\\
Bearing1-7     & \textbf{0.108} & \textbf{0.022} & 0.224 & 0.1854 & \textbf{100.0} & 0.2038 & 0.1635 & 99.8668\\
\hline
Bearing2-3     & 0.131 & 0.033 & 0.1306 & 0.1012 & \textbf{98.3361} & \textbf{0.1288} & \textbf{0.0993} & 98.9185\\
Bearing2-4     & 0.204 & \textbf{0.081} & \textbf{0.1579} & 0.1295 & 96.732 & 0.1669 & 0.1374 & \textbf{97.7124}\\
Bearing2-5     & 0.202 & \textbf{0.071} & \textbf{0.1319} & 0.116 & 88.2617 & 0.1523 &  0.1311 & \textbf{94.955} \\
Bearing2-6     & 0.205 & \textbf{0.083} & \textbf{0.2167} & 0.1566 & \textbf{100.0} & 0.2275 &  0.1739 & \textbf{100.0}\\
Bearing2-7     & 0.397  & 0.220 & 0.1398 & 0.1113 & \textbf{100.0} & \textbf{0.1391} &  \textbf{0.1082} & \textbf{100.0} \\
\hline
Bearing3-3     & 0.280          & 0.161 & \textbf{0.2142} & \textbf{0.1097} & \textbf{100.0} & 0.2163 & 0.1125 & 93.75\\
\hline
\end{tabular}
}
\label{Pronostia_dependent}
\end{table}

Tables \ref{XJTU_dependent} and \ref{Pronostia_dependent} show the performance analysis of our model using the OC-dependent splitting rule. It is worth noting that only SACGNet was considered for the analysis because the other models did not utilize the OC-dependent rule. Our proposed models showed significant superiority over the SACGNet model for all bearings of the XJTU-SY dataset. In the PRONOSTIA dataset, our models performed notably better than SACGNet in almost all bearings, except for $Bearing1_{-}5$ and $Bearing1_{-}7$ in terms of RMSE. Our proposed model demonstrated competitive performance compared to the SACGNet model regarding MAE scores in the PRONOSTIA dataset. It is worth noting that the OC classification of $Bearing1_{-}4$ in Table \ref{XJTU_dependent} was relatively poor. The poor performance of this bearing can be attributed to its unique features, which significantly differ from other bearings operating under the same conditions. This observation has been reported in related works~\cite{huang2021novel}. Finally, the obtained results underscore again the significant improvements in RMSE and MAE scores across almost all bearing types when the denoised LSTM-Autoencoder is used. 
\section{Conclusion}\label{conclusion}
This paper presented the robust MDL model for the prediction of Remaining Useful Life (RUL) and the classification of Operating Conditions (OC) of rotating machines. The model comprises several key components: a denoising LSTM-autoencoder responsible for data denoising, three parallel branches (1D data branch, 2D data branch based on Resnet-34 architecture, and a denoised data branch) for feature extraction, AB-LSTM blocks for RUL prediction, and GAP blocks for OC classification. This parallel architecture empowers the proposed model to capture intricate relationships between bearing operating conditions and degradation patterns, resulting in superior performance in both RUL prediction and OC classification tasks. Furthermore, in addition to the raw data, a comprehensive set of features, including 11 time-domain, 3 frequency-domain, and 2D time-frequency domain features, is computed and utilized as rich input for our model. To assess the model's performance, we compared it to state-of-the-art models on both the PRONOSTIA and XJTU-SY datasets. The obtained results indicate that our model outperforms others on both datasets, making it a promising option for predictive maintenance applications. Utilizing the LSTM-Autoencoder for data denoising is a crucial step towards enhancing the robustness of the model. Its application leads to a significant improvement in the overall performance and accuracy. In our future work, we aim to test the robustness and performance of our models on real applications. We also plan to extend the models to incorporate additional data sources, such as expert opinions and machine sounds. 
\bibliographystyle{plain}
\bibliography{cas-refs}
\end{document}